\documentclass[11pt,letter]{article}

\usepackage{amssymb}
\usepackage{amsmath}
\usepackage{epsfig}
\usepackage{graphicx}
\usepackage{color}
\usepackage{bigints}
\usepackage{array}
\usepackage[normalem]{ulem}
\usepackage{xspace}
\usepackage{algorithm}
\usepackage{algorithmic}
\usepackage{enumerate}
\usepackage{hyperref}
\usepackage{numprint}
\usepackage{subcaption}
\usepackage{changepage}
\usepackage{enumerate}
\usepackage{wrapfig}

\usepackage{amsmath}
\usepackage{amssymb}
\usepackage{mathtools}
\usepackage{amsthm}

\usepackage{multirow}
\usepackage{amssymb}
\usepackage{amsmath}
\usepackage{epsfig}
\usepackage{graphicx}
\usepackage{color}
\usepackage{bigints}
\usepackage{array}
\usepackage[normalem]{ulem}
\usepackage{xspace}
\usepackage{algorithm}
\usepackage{algorithmic}
\usepackage{enumerate}
\usepackage{hyperref}
\usepackage{numprint}
\usepackage{subcaption}
\usepackage{changepage}
\usepackage{enumerate}
\usepackage{wrapfig}

\graphicspath{{./}{figures/}}

\newcommand{\real}{\mathbb{R}}

\newcommand{\eps}{\varepsilon}
\newcommand{\mat}[1] {\mathtt{#1}}
\newcommand{\supp} {\mathtt{supp}}

\newcommand{\cancel}[1] {}

\usepackage{amsthm}

\newtheorem{lemma}{Lemma}[section]

\newtheorem*{corollary*}{Corollary}

\newbox\ProofSym
\setbox\ProofSym=\hbox{%
\unitlength=0.18ex%
\begin{picture}(10,10)
\put(0,0){\framebox(9,9){}}
\put(0,3){\framebox(6,6){}}
\end{picture}}

\begin{document}

\begin{titlepage}

\title{A Fast Method for Lasso and Logistic Lasso\thanks{Research supported by Research Grants Council, Hong Kong, China (project no.~16203718).}}

\author{Siu-Wing Cheng\footnote{Department~of~Computer~Science~and~Engineering,
		HKUST, Hong Kong. Email: {\tt scheng@cse.ust.hk}, {\tt mtwongaf@connect.ust.hk}}
	\and 
	Man Ting Wong\footnotemark[2]}

\date{}

\maketitle
 
\begin{abstract}
We propose a fast method for solving compressed sensing, Lasso regression, and Logistic Lasso regression problems that iteratively runs an appropriate solver using an active set approach. 
We design a strategy to update the active set that achieves a large speedup over a single call of several solvers, including gradient projection for sparse reconstruction (GPSR), lassoglm of Matlab, and glmnet.  For compressed sensing, the hybrid of our method and GPSR
is 31.41 times faster than GPSR on average for Gaussian ensembles and 25.64 faster on average for binary ensembles.  For Lasso regression, the hybrid of our method and GPSR achieves a 30.67-fold average speedup in our experiments. In our experiments on Logistic Lasso regression, the hybrid of our method and lassoglm gives an 11.95-fold average speedup,
and the hybrid of our method and glmnet 
gives a 1.40-fold average speedup.
\end{abstract}

\end{titlepage}

\section{Introduction}

\paragraph{Lasso Regression.}  
Lasso~\cite{lasso} is designed as an extension of the linear regression.  Consider a phenomenon that involves $\nu$ explanatory variables.  Suppose that there is a dataset of $n$ observations, each specifying the values of these $\nu$ explanatory variables and the value of the corresponding response which are real numbers.  This dataset can be captured by a matrix $\mat{A} \in \mathbb{R}^{n \times \nu}$ with one column for each explanatory variable and a vector $\mat{b} \in \mathbb{R}^n$ with one coordinate for each response.
Lasso fits a model $\mat{x} \in \mathbb{R}^\nu$ by solving the following problem:

$$\min \|\mat A\mat x-\mat b\|^2_2 \quad \text{subject to} \quad \|\mat x\|_1 \le t,$$
where $\|\mat x\|_1$ is the $L_1$-norm of $\mat{x}$.  Each coordinate of $\mat{x}$ can be positive, zero, or negative. We call it the \emph{constrained form} of Lasso.  Alternatively, for a  coefficient $\eta > 0$ chosen in advance, the problem can be written as:
$$\min \|\mat A\mat x-\mat b\|^2_2 + \eta \|\mat x \|_1.$$
 We call it the \emph{Lagrangian form} of Lasso.

Lasso can be used for linear regression and variable selection. It was used to predict the level of prostate-specific antigen \cite{lasso}, neighborhood selection for sparse high-dimensional graphs \cite{hdgv}, variable selection for corporate bankruptcy forecasts \cite{vscb}, predicting high-growth firms \cite{cglb}, and prioritizing driving factors of household carbon emissions \cite{pdfh}.

Lasso is closely related to, and has been applied to basis pursuit problems in signal processing \cite{ADbp}. The basis pursuit is defined as $\min \|\mat x\|_1\, s.t.\  \|\mat A\mat x-\mat b\|^2_2\le \eps$. Basis pursuit can also be considered in its Lagrangian form: $\min \|\mat A\mat x-\mat b\|^2_2+ \eta \|\mat x\|_1$ \cite{gpsr}.

The Lasso problem uses the $L_1$-norm to help find sparse solutions, where a  solution is sparse if it has a small number of non-zero entries. As mentioned in \cite{gpsr}, Lasso was used in several signal processing problems where sparse solutions are desired \cite{rmed,rsst,dwtn}. Lasso can also be applied to wavelet-based image/signal reconstruction and restoration (deconvolution) \cite{idws,eawb,nosr}. It is also applied in compressed sensing, where the original signals can be reconstructed from a substantially smaller number of observations, provided that some prior criteria are met for the measurement matrix \cite{ssri,nosr,dsse,rupe,cs}.

In this paper, we will focus on the Lagrangian form of Lasso.

\paragraph{Logistic Lasso Regression.} 
In Logistic Lasso, a response is either $0$ or $1$, and the goal is to predict the probability of the response being $1$ for given values of the explanatory variables.  This probability is formulated as $1 / (1+\exp{( -\mat{a}^t \mat{x})})$, where $\mat{a}$ is the vector consisting of a $1$ followed by the given values of the explanatory variables, and $\mat{x} \in \mathbb{R}^\nu$ is the unknown model to be determined~\cite{glmlassopaper}.

The model $\mat{x}$ is estimated using $n$ observations that specify the values of the explanatory variables together with the responses.  This gives $n$ vectors $\mat{a}_i$ and $n$ binary values $y_i$ for $i \in [n]$.  The probability of $y_i$ being 1 is $\mu_i = 1 / (1+\exp{( -\mat{a}_i^t \mat{x})})$.  Hence, $\mat{x}$ can be determined by minimizing the negative log-likelihood function~\cite{mlpp, glmlassopaper}.  That is,
$$\begin{array}{c}
\displaystyle \min -\sum^n_{i=1}  \bigl(y_i \log \mu_i + (1-y_i) \log(1-\mu_i)\bigr) \\[.5em]
\text{subject to} \|\mat x\|_1 \le t,
\end{array}$$
where $t > 0$ is fixed \emph{a priori}.  The Logistic Lasso regression is also considered in its Lagrangian form: 
$$\min -\frac{1}{n}\sum^n_{i=1}  \bigl(y_i \log \mu_i + (1-y_i) \log(1-\mu_i)\bigr) +\eta \|\mat x\|_1,$$
where $\eta>0$ is fixed beforehand, and the term $\frac{1}{n}$ is included at the beginning for convention. 

The Logistic Lasso regression is used for binary classification and variable selections; for example, diagnosis of atypical Crohn's disease~\cite{crohn}, genome-wide association analysis~\cite{genome}, and credit scoring problems~\cite{cscoring}. 

In this paper, we will focus on the Lagrangian form of Logistic Lasso.

\paragraph{Choice of solvers.} Our algorithm iteratively runs a solver using an active set approach, so we need to adopt some appropriate solvers.

For compressed sensing and Lasso regression, it has been reported in~\cite{gpsr} that gradient projection for sparse reconstruction (GPSR) is more efficient than several other solvers, including IST~\cite{dd04, eawb}, l1\_ls by Boyd and Lustig et al.~\cite{kse07},
the Homotopy method~\cite{ddy08}, and l1-magic~\cite{l1magic}. GPSR has garnered significant attention in the research community, with 2,428 citations in papers and 41 citations in patents according to IEEE Xplore
and 4,316 citations according to Google Scholar. We use the GPSR-BB version, which we refer to as GPSR in this paper.

For compressed sensing, another popular method is the alternating direction method of multipliers (ADMM)~\cite{ADMM_O1,ADMM_O2}. 
The review of ADMM~\cite{ADMM_re} has been cited 21,656 times.  This solver is available on Boyd's website \cite{ADMM_web}.

The solver {glmnet} is a strong contender for Lasso and Logistic Lasso regressions. The underlying algorithm was developed by Tibshirani et al.~\cite{glmlassopaper}. It has been reported~\cite{glmlassopaper} that the algorithm is more efficient than {lars}~\cite{lars} for Lasso and several solvers for Logistic Lasso, including {l1logreg}~\cite{logreg}, {BBR}~\cite{bbr}, and {LPL}~\cite{LPL}. The algorithm has 5,528 citations as reported by the National Library of Medicine and 16,824 citations according to Google Scholar.

The solver {lassoglm} shares the same underlying algorithm with {glmnet}. Since citations of solvers typically point to the underlying algorithm,
distinguishing the popularity of {lassoglm} and {glmnet} can be challenging. Thus, we also experimented with {lassoglm}.
It is a built-in function within a Matlab toolbox, so it is easily accessible.

We study the hybrids of our method with GPSR, ADMM, lassoglm, and glmnet in our experiments.  For compressed sensing, we use GPSR, ADMM, and lassoglm.  For Lasso regression, we use GPSR, lassoglm and glmnet.  For Logistic Lasso regression, we use lassoglm and glmnet. The URLs to these solvers are listed in the appendix.

\paragraph{Our contributions.} We propose a fast method for Lasso or Logistic Lasso that iteratively runs an appropriate solver using an active set approach.  Compressed sensing is covered as well because it is a variant of Lasso.  

The essence is to put many variables into the active set and hence keep them at zero; this makes a run of the solver much more efficient.  The difficulty lies in freeing variables from the active set from one iteration to the next.  If all variables are freed, the optimum will be reached in one go, but efficiency suffers because the solver needs to optimize over many variables.  In previous works, researchers tried to free the variable that causes the ``most violation'' of the optimality condition or a constant number of variables that do so.  This strategy can lead to many iterations which is not desirable.

In~\cite{nnqp}, it is discovered that one can obtain large speedups by freeing $\Theta(\log^2 \nu)$ variables from the active set for non-negative convex quadratic programming, provided that the intermediate problems have sparse solutions.  Variables at zero are moved into the active set.  The function $\Theta(\log^2 \nu)$ scales with $\nu$, yet it 
is small enough that the solver does not suffer from dealing with many variables.

We adapt this idea to Lasso and Logistic Lasso because the solutions are often sparse in such applications.   However, the adaptation is far from straightforward because we need to deal with the non-smoothness of the objective function.  We design a novel strategy for updating the active set that has very different theoretical underpinnings from that in~\cite{nnqp}.  We show that under a heuristical assumption, freeing $O(\log^4 \nu)$ variables from the active set ensures a descent direction that is within a constant angle from the direction of the optimum.  Experimentally, by freeing $\Theta(\log^2 \nu)$ variables from the active set, the hybrids of our method with GPSR, ADMM, lassoglm, and glmnet produce significant performance gains.  We achieve a 31.41-fold average speedup for GPSR for compressed sensing with Gaussian ensembles, a 25.64-fold average speedup for GPSR for compressed sensing with binary ensembles, an 80-fold or more average speedup for ADMM for compressed sensing, a 30.67-fold average speedup for GPSR for Lasso linear regression, an 11.92-fold average speedup for lassoglm for Logistic Lasso, and a 1.40-fold average speedup for glmnet for Logistic Lasso.

\section{Algorithm}
\label{sec:alg}

\paragraph{Notation.} 
Matrices and vectors are represented by uppercase and lowercase letters in typewriter font, respectively. We denote the inner product of vectors $\mat{x}$ and $\mat{y}$ as $\langle \mat{x},\mat{y} \rangle$ or $\mat{x}^t\mat{y}$. The notation $0_m$ stands for the $m$-dimensional zero vector, and $1_m$ stands for the $m$-dimensional vector with all coordinates equal to 1.

For $i \in [\nu]$, $\mat{e}_i$ denotes the unit vector on the $i$-th positive coordinate axis. The $i$-th coordinate of $\mat{e}_i$ is 1 and all other coordinates of $\mat{e}_i$ are zeros. A \emph{conical combination} of a set $W$ of vectors is $\sum_{\mat{w} \in W} c_\mat{w} \mat{w}$ for some non-negative real coefficients $c_\mat{w}$'s. For instance, the set of all conical combinations of ${\mat{e}_1,\mat{e}_2,\ldots,\mat{e}_\nu}$ is the positive quadrant in $\real^\nu$.

 We use $(\mat{x})_i$ to denote the $i$-th coordinate of a vector $\mat{x}$. 
 It is not to be confused with a vector that is labeled with a subscript. For instance, we have $(\mat{e}_i)_i = 1$ and $(\mat{e}_i)_j = 0$ whenever $j \neq i$. In our algorithm, a Lasso solver is invoked iteratively, thereby producing a sequence of intermediate solutions. We use $\mat{x}_r$ to represent the intermediate solution that is computed in the $(r-1)$-th round.  The $i$-th coordinate of $\mat{x}_r$ is $(\mat{x}_r)_i$. 
We use $\mbox{supp}(\mat{x})$ to denote $\bigl\{i : (\mat{x})_i \neq 0 \bigr\}$.

\paragraph{Duality and Slater's condition.}  Let the objective function be denoted by $F: \real^\nu \rightarrow \real$.  We have $F(\mat{x}) = f(\mat{x}) + \eta\|\mat{x}\|_1$, where $f(\mat{x})$ is $\|\mat{Ax}-\mat{b}\|_2^2$ for Lasso; and  $f(\mat{x})$ is  $-\frac{1}{n}\sum^n_{i=1}  \bigl(y_i \log \mu_i + (1-y_i) \log(1-\mu_i)\bigr)$ for Logistic Lasso.

For $r = 2, 3, \ldots$, in the $(r-1)$-th iteration of our algorithm, there is an active set $S_r$ that stores the indices of some variables that are kept at zero, and a solver is run to solve the Lagrangian form of the Lasso or Logistic Lasso problem under this constraint.
We say that the variables whose indices belong to $S_r$ are the variables in the active set $S_r$.  We say that the other variables are \emph{free}.

We are interested in the dual problem~\cite{BV2004} as it tells us how to update the active set $S_r$ when proceeding to the next iteration.

For the constraint $(\mat{x})_i = 0$ induced by $i \in S_r$, there is a Lagrange multiplier that is a variable in the dual problem, and it can take on any real value.  For convenience, we also create a multiplier and variable in the dual problem for each $i \not\in S_r$, and we fix these variables at zero.  In all, we have a vector $\mat{v}$ of variables for the dual problem such that $(\mat{v})_i$ can be any real number for $i \in S_r$ and $(\mat{v})_i = 0$ for $i \not\in S_r$.



The dual problem has $F(\mat x) -  \sum_{j \in S_r} (\mat v)_j \cdot (\mat x)_j$ as its objective function~\cite{BV2004}.  By our design of $\mat{v}$, we can equivalently express this objective function as $G(\mat{v},\mat{x}) = F(\mat{x}) - \mat{v}^t\mat{x}$. Figure~\ref{fg:NNM-main} gives the primal and dual problems.

\begin{figure}[h]
\centering
\begin{tabular}{ccc}
$\min \,\, F(\mat{x})$ && $\displaystyle \max_{\mat{v}} \,\, \min_{\mat{x}} \,\, G( \mat{v},\mat{x})$ \\[.5em] 
$\forall \, i \in S_r, \, (\mat{x})_i = 0$ && 
$\forall \, i \not\in S_r, \,(\mat{v})_i = 0$. \\[1em]
(a)~Lagrangian form && (b)~dual problem \\
\end{tabular}
\caption{The Lagrangian form of Lasso/Logistic Lasso and its dual problem.}
\label{fg:NNM-main}
\end{figure}

It is known that the optimal value of the dual problem is a lower bound of the optimal value of the primal problem.
This is known as the \emph{weak duality}.
Under some situations, \emph{strong duality} holds which means that the optimal solutions of the primal and dual problems have the same value.  One such sufficient condition is the \emph{Slater's condition} which requires the existence of a feasible point of the primal problem that strictly satisfies every feasibility inequality constraint~\cite{BV2004}.

In our case, there is no inequality constraint in the primal problem, and the origin is always a feasible solution to the primal problem no matter what $S_r$ is.  Slater's condition is thus satisfied which guarantees strong duality.

\paragraph{Subgradient.} 

A vector $\mat{g} \in \mathbb{R}^\nu$ such that $F(\mat y) \ge F(\mat x) + \langle \mat{g}, \mat y - \mat x \rangle$ for all $\mat y \in \mathbb{R}^\nu$ is called a \emph{subgradient} of $F$ at $\mat{x}$~\cite{nonl}.  For a smooth function, the subgradient at a point is unique and equal to the gradient.  This is not so for $F$ as it is non-smooth.

Let $\partial F(\mat{x})$ denote the set of all subgradients of $F$ at $\mat{x}$.  Note that $\partial F(\mat x)\neq \emptyset$ for every feasible point $\mat{x}$.  Recall that $F(\mat{x}) = f(\mat{x}) + \eta\|\mat{x}\|_1$.

A useful way to visualize subgradients is as follows.  The plot of $(\mat{x},F(\mat{x}))$ is a convex hypersurface $\mathcal{C}$ in $\real^{\nu+1}$.  Take a particular point $(\mat{z},F(\mat{z}))$ on $\mathcal{C}$.  Let $\mat{g}$ be a subgradient at $\mat{z}$.  Consider the hyperplane $H$ in $\real^{\nu+1}$ through $(\mat{z},F(\mat{z}))$ that is orthogonal to the vector $(\mat{g}^t,-1)^t$.  By the subgradient definition, $F(\mat{x}) \geq F(\mat{z}) + \langle \mat{g}, \mat{x}-\mat{z} \rangle$ for all $\mat{x} \in \real^\nu$, which is equivalent to $\mathcal{C}$ lying above $H$.  In other words, $H$ is tangent to $\mathcal{C}$ at $\mat{z}$.  Conversely, each hyperplane $H$ tangent to $\mathcal{C}$ induces a unique vector $(\mat{g}^t,-1)^t$ orthogonal to $H$, and one can easily verify that $\mat{g}$ is a subgradient at the projection of the tangential contact point between $H$ and $\mathcal{C}$ to $\mathbb{R}^\nu$.  

Take a point $\mat{z} \in \real^\nu$.   Let $I = \{i : (\mat{z})_i = 0\}$, let $I^+ = \{i : (\mat{z})_i > 0\}$, and let $I^- = \{i : (\mat{z})_i < 0\}$.   Consider the case that $I = \emptyset$.  Then, $F$ is smooth at $\mat{z}$.  So the gradient $\nabla F(\mat{z})$ is defined, and $\nabla F(\mat{z})$ is the unique subgradient at $\mat{z}$.  In this case, if $i \in I^+$, then $\frac{\partial F(\mat{z})}{\partial (\mat{x})_i} = \frac{\partial f(\mat{z})}{\partial (\mat{x})_i} + \eta$, and if $i \in I^-$, then $\frac{\partial F(\mat{z})}{\partial (\mat{x})_i} = \frac{\partial f(\mat{z})}{\partial (\mat{x})_i} - \eta$. Consider the case that $I \not= \emptyset$.  There are $2^{|I|}$ different ways to move slightly from $\mat{z}$ to make the $i$-th coordinate positive or negative.
\begin{itemize}

    \item There are $2^{|I|}$ tuples $(s_i)_{i \in I} \in \{-1,1\}^{|I|}$. There is one subgradient $\mat{g}_\sigma$ for each tuple $\sigma = (s_i)_{i \in I}$ such that $(\mat{g}_\sigma)_i = \frac{\partial f(\mat{z})}{\partial (\mat{x})_i} + \eta$ for $i \in I^+$, $(\mat{g}_\sigma)_i = \frac{\partial f(\mat{z})}{\partial (\mat{x})_i} - \eta$ for $i \in I^-$, and $(\mat{g}_\sigma)_i = \frac{\partial f(\mat{z})}{\partial (\mat{x})_i} + s_i\eta$ for $i \in I$.  

    \item From the previous discussion of the relation between subgradient and tangent hyperplanes of $\mathcal{C}$, we know that for every subgradient $\mat{g}$ at $\mat{z}$, $(\mat{g}^t,-1)^t$ is a convex combination of the $(\mat{g}_\sigma^t,-1)^t$'s over all $2^{|I|}$ tuples $\sigma$ of $\{-1,1\}^{|I|}$.
    
\end{itemize}
By convex combinations, for each $i \in I$, any value between $\frac{\partial f(\mat{z})}{\partial (\mat{x})_i} - \eta$ and $\frac{\partial f(\mat{z})}{\partial (\mat{x})_i} + \eta$ can be generated independently as the $i$-th coordinate of a subgradient.  This can be seen as follows.  

Take any $j \in I$.  Take two tuples $\sigma_1, \sigma_2 \in \{-1,1\}^{|I|}$ that differ only in $s_j$.  We can choose coefficients for $\mat{g}_{\sigma_1}$ and $\mat{g}_{\sigma_2}$ so that their convex combination $\mat{g}$ has the target $j$-th coordinate.  Take another index $k \in I\setminus \{j\}$.  Assume without loss of generality that the $k$-th coordinate of $\mat{g}$ is $\frac{\partial f(\mat{x}_r)}{\partial (\mat{x})_k} + \eta$.  We choose two tuples $\sigma_1',\sigma'_2 \in \{-1,1\}^{|I|}$ that are the same as $\sigma_1$ and $\sigma_2$, respectively, except that their common $s_k$ is $-1$ (instead of 1).  We can combine $\mat{g}_{\sigma'_1}$ and $\mat{g}_{\sigma'_2}$ as before to produce a vector $\mat{g'}$ such that $(\mat{g}')_i = (\mat{g})_i$ for $i \in I\setminus \{k\}$ and $(\mat{g}')_k = \frac{\partial f(\mat{x}_r)}{\partial (\mat{x})_k} - \eta$. Now, we can compute a convex combination of $\mat{g}$ and $\mat{g}'$ that has the desired $k$-th coordinate.  Note that the $j$-th coordinate is unchanged and remains at its target value.  Continuing in this manner, for any $i \in I$, we can set the $i$-th coordinate of the subgradient to any value between $\frac{\partial f(\mat{z})}{\partial (\mat{x})_i} - \eta$ and $\frac{\partial f(\mat{z})}{\partial (\mat{x})_i} + \eta$.

The above discussion leads to the following result.

\vspace{6pt}

\begin{lemma}
\label{lem:subgradient}
   Let $\mat{z}$ be a feasible point of the primal problem.  Let $I = \{i : (\mat{z})_i = 0\}$, let $I^+ = \{i : (\mat{z})_i > 0\}$, and let $I^- = \{i : (\mat{z})_i < 0\}$.  Then, $\mat{g} \in \partial F(\mat{z})$ if and only if:
   \begin{enumerate}[{\em (i)}]
    \item For each $i \in I^+$, $(\mat{g})_i = \frac{\partial f(\mat{z})}{\partial (\mat{x})_i} + \eta$.  
    
    \item For each $i \in I^-$, $(\mat{g})_i = \frac{\partial f(\mat{z})}{\partial (\mat{x})_i} - \eta$.   
    
    \item For each $i \in I$, $(\mat{g})_i \in \Bigl[\frac{\partial f(\mat{z})}{\partial (\mat{x})_i}-\eta,\frac{\partial f(\mat{z})}{\partial (\mat{x})_i}+\eta\Bigr]$.
   \end{enumerate}
\end{lemma}

\paragraph{KKT conditions.} Our algorithm runs iteratively. Let $r\ge2$ be an integer and let $\mat{x}_r$ denote the optimal solution of the constrained Lasso problem in Figure~\ref{fg:NNM-main}(a).
Let $(\mat{v}_r,\mat{x}_r)$ be the corresponding solution of the dual problem.  We call $\mat{x}_r$ an optimal primal solution constrained by $S_r$ and $(\mat{v}_r,\mat{x}_r)$ an optimal dual solution constrained by $S_r$.  

The Karush-Kuhn-Tucker conditions, or KKT conditions for short, are the sufficient and necessary conditions for $(\mat{v}_r,\mat{x}_r)$ to be an optimal dual solution~\cite{nonl}.  There are four KKT conditions:
\begin{itemize}
	\item Critical point: $\mat{v}_r \in \partial F(\mat{x}_r)$.
	
	\item Primal feasibility:  $\forall \, i \in S_r$, $(\mat{x}_r)_i = 0$.
	
	\item Dual feasibility:  $\forall\, i \not\in S_r $, $(\mat{v}_r)_i = 0$.
	
	\item Complementary slackness:  $\forall \, i$, $(\mat{v}_r)_i \cdot (\mat{x}_r)_i = 0$.

\end{itemize}

Complementary slackness is implied by the primal and dual feasibilities.  Therefore, the KKT conditions boil down to the existence of $\mat{x}_r \in \real^\nu$ and a subgradient $\mat{v}_r \in \partial F(\mat{x}_r)$ that satisfy the primal and dual feasibilities.

The dual solution $(\mat{v}_r,\mat{x}_r)$ satisfies the KKT conditions.  To proceed to the next iteration, we should find an index $j \in S_r$ such that removing $j$ from $S_r$ leads to a violation of the KKT conditions.  As a result, freeing the variable $(\mat{x})_j$ will allow the objective function value to decrease further.  Lemma~\ref{lem:rule} below tells us how to find such an index.

\begin{lemma}
    \label{lem:rule}
    The KKT conditions are violated after removing an index $j$ from $S_r$ if and only if $\bigl|\frac{\partial f(\mat{x}_r)}{\partial (\mat{x})_j}\bigr| > \eta$.
\end{lemma}
\begin{proof}
From our previous discussion, removing $j \in S_r$ violates the KKT conditions
if and only if there does not exist a subgradient $\mat{w} \in \partial F(\mat{x}_r)$ such that $(\mat{w})_i = 0$ for all $i \not\in S_r\setminus\{j\}$.  Let $I_r = \{i : (\mat{x}_r)_i = 0\}$.
We have $j \in I_r$ as $S_r \subseteq I_r$.  By Lemma~\ref{lem:subgradient}, the $j$-th coordinates of the subgradients at $\mat{x}_r$ form the interval $\bigl[\frac{\partial f(\mat{x}_r)}{\partial (\mat{x})_j}-\eta,\frac{\partial f(\mat{x}_r)}{\partial (\mat{x})_j}+\eta\bigr]$.  Therefore, $0 \in \bigl[\frac{\partial f(\mat{x}_r)}{\partial (\mat{x})_j}-\eta,\frac{\partial f(\mat{x}_r)}{\partial (\mat{x})_j}+\eta\bigr]$ if and only if $\bigl|\frac{\partial f(\mat{x}_r)}{\partial (\mat{x})_j}\bigr| \leq \eta$.  In other words, removing $j$ from $S_r$ violates the KKT conditions if and only if $\bigl|\frac{\partial f(\mat{x}_r)}{\partial (\mat{x})_j}\bigr| > \eta$.
\end{proof}

\vspace{-4pt}

\paragraph{Algorithm.}  A variable $(\mat{x})_j$ in $S_r$ is \emph{eligible} if $\bigl|\frac{\partial f(\mat{x}_r)}{\partial (\mat{x})_j}\bigr| > \eta$.  Given two eligible variables $(\mat{x})_j$ and $(\mat{x})_k$ in $S_r$, we say that $(\mat{x})_j$ is \emph{larger} than $(\mat{x})_k$ if $\bigl|\frac{\partial f(\mat{x}_r)}{\partial (\mat{x})_j}\bigr| > \bigl|\frac{\partial f(\mat{x}_r)}{\partial (\mat{x})_k}\bigr|$.

Our algorithm uses a parameter $\tau > 0$.  If there are at least $\tau$ eligible variables in $S_r$, we free the $\tau$ largest ones.  If there are fewer than $\tau$ eligible variables in $S_r$, we free them all.  We set $\tau = \lfloor 4\ln^2 \nu \rfloor$ in our experiments, and we did \emph{not} optimize the setting of $\tau$ for different datasets. There are $\nu$ variables in the whole system, so $\tau$ is only a tiny fraction.  We will explain later why such a small $\tau$ should work.
 
Algorithm~\ref{alg:1} gives the pseudocode of our algorithm.  Lines~\ref{alg:tau} and~\ref{alg:update} are the key steps for updating the active set.  It means that there will be $|\supp(\mat{x}_r)| + \tau$ free variables only in the next iteration.  For problems with sparse intermediate solutions (i.e., $|\supp(\mat{x}_r)| \ll \nu$), each iteration will run fast.

\begin{algorithm}
\caption{SolveLasso}\label{alg:1}
\begin{algorithmic}[1]
        \STATE Initialize $S_1$.  
        \STATE Find the optimal solution $(\mat{v}_1,\mat{x}_1)$ of the dual problem constrained by $S_1$.
	\STATE $r \leftarrow 1$
	\WHILE{there is some eligible variable in $S_r$}
		\STATE{$E_r \leftarrow \text{the ordered list of eligible variables in $S_r$}$}
	    \IF {$|E_r| \ge \beta_0$ and $r \le \beta_1$}   					\STATE{$E'_r \leftarrow \text{the largest $\tau$ eligible variables in $E_r$}$}  \label{alg:tau}
		      \STATE{$S_{r+1} \leftarrow [\nu] \setminus \bigl(\supp(\mat{x}_r) \cup E'_r\bigr)$}  \label{alg:update}
	    \ELSE
		      \STATE $S_{r+1} \leftarrow S_r \setminus E_r$				 
	    \ENDIF
	    \STATE Find the optimal solution $(\mat{v}_{r+1},\mat x_{r+1})$ of the dual problem constrained by $S_{r+1}$. \label{alg:qp}
	    \STATE $r \leftarrow r + 1$
        \ENDWHILE
        \STATE \textbf{return} $\mat{x}_r$ 
\end{algorithmic}
\end{algorithm}

There are two other parameters $\beta_0$ and $\beta_1$ in addition to $\tau$. The parameter $\beta_0$ should be $c\tau$ for some constant $c > 1$, and we set $\beta_0 = 3\tau$ in our experiments.  If $|E_r| \leq \beta_0$, it is safer to free all eligible variables in the active set than risk any erroneous update of the active set because the number of free variables in the next iteration will not be too large anyway.  The parameter $\beta_1$ is an upper bound on the number of iterations.  From our computational experience, the objective function value drops at least geometrically, so a small constant for $\beta_1$ should allow the algorithm to reach the optimum ($\beta_1 = 15$ in our experiments).  Something must have gone very wrong in the rare event that $\beta_1$ is exceeded.  As a fail-safe measure, when $\beta_1$ is exceeded, we free all eligible variables in the active set to ensure convergence.

At a high level, Algorithm~\ref{alg:1} is similar to the algorithm in~\cite{nnqp} for non-negative convex quadratic programming.  However, our active set update rule---the key step of Algorithm~\ref{alg:1}---is new and has very different theoretical underpinnings as stated in Lemmas~\ref{lem:subgradient} and~\ref{lem:rule}.

\paragraph{Why does a small $\tau$ work?}  

Let $\mat{d}$ be the descent direction from the current solution towards the optimum.  We show that under a heuristical assumption about $\mat{d}$, $O(\log^4 \nu)$ eligible variables can induce a conical combination $\mat{u}$ such that $\angle (\mat{d},\mat{u}) < 64^\circ$.  We can thus descend in an ``approximately correct'' direction unless there are fewer than $\beta_0$ eligible variables, or  the algorithm has iterated more than $\beta_1$ times. In the following, we assume that there are $\Omega(\log^4 \nu)$ eligible variables.  .

Each eligible variable corresponds to a coordinate axis in $\real^\nu$.  Let $\mat{e}_1,\ldots,\mat{e}_m$ be the unit vectors on the positive coordinate axes for the eligible variables.  Let $X$ be the set of points including the origin, $\mat{d}$, $\mat{e}_1,\ldots,\mat{e}_m$.  Take a random projection $\varphi$ of $X$ to an Euclidean space of dimension $s = \kappa\varepsilon^{-2}\log^2 m$ for some appropriate constant $\kappa > 0$.  It holds with high probability that the distances between the points in $X$ are preserved with distortion $\leq 1+\varepsilon$, and so is the volume of any simplex formed by at most $2 + \log m$ points in $X$~\cite{M2007}.  That is, the projection does not change them by a factor more than $1+\varepsilon$ with high probability.  In particular, the height of any point in $X$ from any affine space spanned by at most $1 + \log m$ points in $X$ is preserved with distortion $\leq 1+\varepsilon$ with high probability.

The direction of $\varphi(\mat{d})$ varies depending on the
location of the current solution and the random projection $\varphi$.  We make the heuristical assumption that the direction of $\varphi(\mat{d})$ is uniformly distributed over the unit sphere in $\real^s$.  

Let $c = \kappa\varepsilon^{-2}$.  Take a hyperplane $H$ in $\mathbb{R}^s$ normal to the vector $\varphi(\mat{e}_1)$, and place $H$ between $\varphi(0)$ and $\varphi(\mat{e}_1)$ at distance $\sqrt{c/s}$ from $\varphi(0)$.  The hyperplane $H$ cuts a cap off the unit sphere that lies on the side of $H$ opposite to $\varphi(0)$.  The ratio of the area of this cap to the area of the unit sphere is $\frac{1}{2}I$, where $I = \int_0^{1-c/s} t^{(s-3)/2} (1-t)^{-1/2}\, dt$~\cite{L2011}. This is also the probability that the angle between the vectors $\varphi(\mat{d})$ and $\varphi(\mat{e}_1)$ is at most $\arccos(\sqrt{c/s})$.  We lower bound $t^{(s-3)/2}(1-t)^{-1/2}$ by $t^{s/2} \geq (1-c/s)^{s/2} \geq (2e)^{-c/2}$, and obtain $I \geq (2e)^{-c/2} \cdot (1-c/s) > \frac{1}{2}(2e)^{-c/2}$.  In all, it holds with probability more than $\frac{1}{4}(2e)^{-c/2}$ that the angle between the vectors $\varphi(\mat{d})$ and $\varphi(\mat{e}_1)$ is at most $\arccos(\sqrt{c/s}) = \arccos(1/\log m)$.  The same argument applies to the other $\varphi(\mat{e}_i)$'s.

Let $\rho = \Theta(\log^4 m)$. 
 Among $\varphi(\mat{e}_1), \ldots, \varphi(\mat{e}_{4(2e)^{c/2}\rho})$, there are $\rho$ of them in expectation that are at an angle no more than $\arccos(1/\log m)$ from the vector $\varphi(\mat{d})$.  Let these $\rho$ vectors be $\varphi(\mat{e}_1),\ldots,\varphi(\mat{e}_\rho)$.  We generalize Lemma~5 in~\cite{nnqp} to show that a conical combination of $\varphi(\mat{e}_1), \ldots, \varphi(\mat{e}_\rho)$ can increase the cosine of the angle by a factor $(4/9) \cdot \log m$.  That is, there is a conical combination $\mat{w}$ of $\varphi(\mat{e}_1), \ldots, \varphi(\mat{e}_\rho)$ that makes an angle at most $\arccos(4/9) = \arcsin(\sqrt{65}/9)$ with $\varphi(\mat{d})$.  The proof of this generalized result is in the appendix.

The conical combination $\mat{w}$ is equal to $\sum_{i=1}^{\rho} \lambda_i \varphi(\mat{e}_i)$ for some $\lambda_i \geq 0$.  If we had included the point $\mat{u} = \sum_{i=1}^{\rho} \lambda_i\mat{e}_i$ in $X$, then $\varphi$ would project $\mat{u}$ to $\mat{w}$ because $\varphi$ is linear and oblivious of the content of $X$.  By the dimension reduction projection, the angle between $\mat{d}$ and $\mat{u}$ would be at most $\arcsin((1+\varepsilon)\sqrt{65}/9) < 64^\circ$ for a small enough $\varepsilon$.  We conclude that there exists a conical combination of $\mat{e}_1,\ldots,\mat{e}_{4(2e)^{c/2}\rho}$ that makes an angle less than $64^\circ$ with $\mat{d}$.  A more precise argument follows the reasoning in the generalized result in the appendix.  The argument above requires $4(2e)^{c/2}\rho =  \Theta(\log^4 \nu)$ eligible variables to be freed.  In practice, freeing the largest $\lfloor 4\log^2 \nu \rfloor$ eligible variables gives very good results.

\section{Experimental results}

In this section, we present our experimental results on compressed sensing, Lasso regression, and Logistic Lasso regression.  Our machine configuration is: Intel Core 7-9700K 3.6Hz cpu, 3600 Mhz ram, 8~cores, 8 logical processors.  We use MATLAB version R2020b. 

We use the GPSR-BB version of the package from~\cite{gpsr} which will be referred to as GPSR.  We also use lassoglm of Matlab, ADMM, and glmnet.  We refer to the hybrids of these solvers with our method as active-GPSR, active-lassoglm, active-ADMM, and active-glmnet.

The solvers GPSR, lassoglm, ADMM, and glmnet take some tolerance parameters that affect the precision of the solutions returned.  Table~\ref{tb:parameter-default} in the appendix shows the default values of these parameters. In some cases (GPSR for compressed sensing and Lasso, and glmnet for Lasso), we run the hybrids with relaxed tolerance parameters to accelerate the solving of the intermediate problems, and upon termination, we run an extra iteration subject to the same final active set but tighter tolerance parameters.  In all other cases, we do not use different tolerance parameters to accelerate the solving of the intermediate problems, so there is no need for an extra iteration.  Table~\ref{tb:parameters} in the appendix gives the corresponding parameter values.

\subsection{Compressed Sensing}

Following part A of the experiment section in \cite{gpsr}, we compare the solvers and their hybrids with our method on recovering a sparse signal.  

The unknown signal is $\mat{z} \in \mathbb{R}^n$.  An ensemble matrix $\mat{A} \in \mathbb{R}^{k \times n}$, where $k < n$, represents a small number of linear probes that are fixed beforehand.  That is, some process gives us a vector $\mat{b}$ which is the product $\mat{Az}$ possibly contaminated with additive Gaussian noise.
The goal is to solve
$\min \frac{1}{2}\|\mat A \mat x - \mat b\|^2_2 +\eta \|\mat x \|_1$ for some fixed $\eta > 0$ and take the solution $\mat{x}$ as an approximation of $\mat{z}$.  Although the least square term has a coefficient of $1/2$ instead of 1, our algorithm is still applicable with no change.

We follow the procedure in~\cite{gpsr} to generate $\mat{A}$, depending on whether $\mat{A}$ is a Gaussian or binary ensemble.  The true signal $\mat{z}$ consists of $s<k$ non-zero entries, each being set randomly to 1 or $-1$.  The $k$ noisy observations are then computed as $\mat{b} = \mat{A} \mat{z} + \mat{n}$, where $\mat{n}$ is the Gaussian white noise, and each entry of $\mat{n}$ has zero mean and variance $10^{-4}$.
%
%
We present our experimental results in the next two subsections.  For additional details on problem generation and selection of regularization parameters, see \cite{gpsr}.

\subsubsection{Gaussian ensemble matrix}

Each matrix entry is sampled identically and independently from the standard Gaussian distribution.  The rows are then orthonormalized to produce the matrix $\mat A$.

The running times with different $n$, $k$, and $s$ are listed in Table \ref{tb:cs1}. Each running time is an average of 10 runs. The first row, with $n=2^{12}$, $k=2^{10}$, and $s=160$, is the configuration used in~\cite{gpsr}. We also try larger values of $n$ and $k$ while keeping the ratio $k/n$ fixed. The number $s$ of non-zeros in the true signal is varied such that $s/n$ is at most the percentage sparsity $160/2^{12}$ in the configuration used in~\cite{gpsr}.

GPSR, ADMM, lassoglm, and glmnet are consistently outperformed by their hybrids with our method.  The speedup of active-GPSR ranges from 3.45-fold to 63.87-fold, with 
a mean of 31.41-fold.   The speedup of active-ADMM ranges from 8.21-fold to 364.51-fold, with 
a mean of 108.10-fold. The speedup of active-lassoglm  ranges from 8.32-fold to 132.90-fold, 
with a mean of 44.50-fold. The speedup of active-glmnet  ranges from 12.40-fold to 131.88-fold, 
with a mean of 52.11-fold.

\begin{table}
\caption{Running time (in seconds) of signal reconstruction using a Gaussian ensemble matrix.  Each running time is an average of 10 trials.}
 \resizebox{\columnwidth}{!}{
\begin{tabular}{|ccccccccccc|}
\hline
\multicolumn{11}{|c|}{Running Time of Signal Reconstruction with Gaussian Ensemble   Matrix}                                                                                                                                                                                                                                                                                       \\ \hline
\multicolumn{1}{|c|}{n}                         & \multicolumn{1}{c|}{k}                         & \multicolumn{1}{c|}{s}    & \multicolumn{1}{c|}{GPSR}  & \multicolumn{1}{c|}{active-GPSR} & \multicolumn{1}{c|}{ADMM}   & \multicolumn{1}{c|}{active-ADMM} & \multicolumn{1}{c|}{lassoglm} & \multicolumn{1}{c|}{active-lassoglm} & \multicolumn{1}{c|}{glmnet} & active-glmnet \\ \hline
\multicolumn{1}{|c|}{$2^{12}$}                  & \multicolumn{1}{c|}{$2^{10}$}                  & \multicolumn{1}{c|}{160}  & \multicolumn{1}{c|}{0.78}  & \multicolumn{1}{c|}{0.23}        & \multicolumn{1}{c|}{3.63}   & \multicolumn{1}{c|}{0.44}        & \multicolumn{1}{c|}{5.07}     & \multicolumn{1}{c|}{0.61}            & \multicolumn{1}{c|}{2.47}   & 0.20          \\ \hline
\multicolumn{1}{|c|}{\multirow{2}{*}{$2^{13}$}} & \multicolumn{1}{c|}{\multirow{2}{*}{$2^{11}$}} & \multicolumn{1}{c|}{160}  & \multicolumn{1}{c|}{1.43}  & \multicolumn{1}{c|}{0.05}        & \multicolumn{1}{c|}{6.92}   & \multicolumn{1}{c|}{0.09}        & \multicolumn{1}{c|}{5.61}     & \multicolumn{1}{c|}{0.18}            & \multicolumn{1}{c|}{4.86}   & 0.17          \\ \cline{3-11} 
\multicolumn{1}{|c|}{}                          & \multicolumn{1}{c|}{}                          & \multicolumn{1}{c|}{320}  & \multicolumn{1}{c|}{3.15}  & \multicolumn{1}{c|}{0.27}        & \multicolumn{1}{c|}{14.85}  & \multicolumn{1}{c|}{1.12}        & \multicolumn{1}{c|}{13.39}    & \multicolumn{1}{c|}{1.05}            & \multicolumn{1}{c|}{8.27}   & 0.50          \\ \hline
\multicolumn{1}{|c|}{\multirow{4}{*}{$2^{14}$}} & \multicolumn{1}{c|}{\multirow{4}{*}{$2^{12}$}} & \multicolumn{1}{c|}{160}  & \multicolumn{1}{c|}{3.43}  & \multicolumn{1}{c|}{0.08}        & \multicolumn{1}{c|}{17.25}  & \multicolumn{1}{c|}{0.09}        & \multicolumn{1}{c|}{10.42}    & \multicolumn{1}{c|}{0.14}            & \multicolumn{1}{c|}{13.37}  & 0.47          \\ \cline{3-11} 
\multicolumn{1}{|c|}{}                          & \multicolumn{1}{c|}{}                          & \multicolumn{1}{c|}{320}  & \multicolumn{1}{c|}{5.38}  & \multicolumn{1}{c|}{0.16}        & \multicolumn{1}{c|}{26.66}  & \multicolumn{1}{c|}{0.26}        & \multicolumn{1}{c|}{18.40}    & \multicolumn{1}{c|}{0.43}            & \multicolumn{1}{c|}{18.47}  & 0.57          \\ \cline{3-11} 
\multicolumn{1}{|c|}{}                          & \multicolumn{1}{c|}{}                          & \multicolumn{1}{c|}{480}  & \multicolumn{1}{c|}{7.72}  & \multicolumn{1}{c|}{0.21}        & \multicolumn{1}{c|}{37.72}  & \multicolumn{1}{c|}{0.55}        & \multicolumn{1}{c|}{25.26}    & \multicolumn{1}{c|}{0.63}            & \multicolumn{1}{c|}{22.84}  & 0.72          \\ \cline{3-11} 
\multicolumn{1}{|c|}{}                          & \multicolumn{1}{c|}{}                          & \multicolumn{1}{c|}{640}  & \multicolumn{1}{c|}{11.73} & \multicolumn{1}{c|}{0.64}        & \multicolumn{1}{c|}{56.48}  & \multicolumn{1}{c|}{1.98}        & \multicolumn{1}{c|}{38.21}    & \multicolumn{1}{c|}{2.03}            & \multicolumn{1}{c|}{31.16}  & 1.50          \\ \hline
\multicolumn{1}{|c|}{\multirow{8}{*}{$2^{15}$}} & \multicolumn{1}{c|}{\multirow{8}{*}{$2^{13}$}} & \multicolumn{1}{c|}{160}  & \multicolumn{1}{c|}{9.00}  & \multicolumn{1}{c|}{0.14}        & \multicolumn{1}{c|}{49.93}  & \multicolumn{1}{c|}{0.14}        & \multicolumn{1}{c|}{24.43}    & \multicolumn{1}{c|}{0.18}            & \multicolumn{1}{c|}{33.58}  & 0.25          \\ \cline{3-11} 
\multicolumn{1}{|c|}{}                          & \multicolumn{1}{c|}{}                          & \multicolumn{1}{c|}{320}  & \multicolumn{1}{c|}{13.05} & \multicolumn{1}{c|}{0.28}        & \multicolumn{1}{c|}{69.54}  & \multicolumn{1}{c|}{0.30}        & \multicolumn{1}{c|}{36.10}    & \multicolumn{1}{c|}{0.45}            & \multicolumn{1}{c|}{48.67}  & 0.41          \\ \cline{3-11} 
\multicolumn{1}{|c|}{}                          & \multicolumn{1}{c|}{}                          & \multicolumn{1}{c|}{480}  & \multicolumn{1}{c|}{17.05} & \multicolumn{1}{c|}{0.41}        & \multicolumn{1}{c|}{88.61}  & \multicolumn{1}{c|}{0.51}        & \multicolumn{1}{c|}{48.71}    & \multicolumn{1}{c|}{0.94}            & \multicolumn{1}{c|}{55.29}  & 0.62          \\ \cline{3-11} 
\multicolumn{1}{|c|}{}                          & \multicolumn{1}{c|}{}                          & \multicolumn{1}{c|}{640}  & \multicolumn{1}{c|}{20.80} & \multicolumn{1}{c|}{0.56}        & \multicolumn{1}{c|}{106.51} & \multicolumn{1}{c|}{0.82}        & \multicolumn{1}{c|}{59.32}    & \multicolumn{1}{c|}{1.27}            & \multicolumn{1}{c|}{66.21}  & 0.84          \\ \cline{3-11} 
\multicolumn{1}{|c|}{}                          & \multicolumn{1}{c|}{}                          & \multicolumn{1}{c|}{800}  & \multicolumn{1}{c|}{24.02} & \multicolumn{1}{c|}{0.64}        & \multicolumn{1}{c|}{122.24} & \multicolumn{1}{c|}{1.14}        & \multicolumn{1}{c|}{67.84}    & \multicolumn{1}{c|}{1.40}            & \multicolumn{1}{c|}{70.46}  & 0.93          \\ \cline{3-11} 
\multicolumn{1}{|c|}{}                          & \multicolumn{1}{c|}{}                          & \multicolumn{1}{c|}{960}  & \multicolumn{1}{c|}{29.05} & \multicolumn{1}{c|}{1.08}        & \multicolumn{1}{c|}{146.81} & \multicolumn{1}{c|}{2.43}        & \multicolumn{1}{c|}{82.70}    & \multicolumn{1}{c|}{2.87}            & \multicolumn{1}{c|}{82.18}  & 1.68          \\ \cline{3-11} 
\multicolumn{1}{|c|}{}                          & \multicolumn{1}{c|}{}                          & \multicolumn{1}{c|}{1120} & \multicolumn{1}{c|}{35.33} & \multicolumn{1}{c|}{1.28}        & \multicolumn{1}{c|}{176.09} & \multicolumn{1}{c|}{3.77}        & \multicolumn{1}{c|}{101.80}   & \multicolumn{1}{c|}{3.54}            & \multicolumn{1}{c|}{94.33}  & 2.25          \\ \cline{3-11} 
\multicolumn{1}{|c|}{}                          & \multicolumn{1}{c|}{}                          & \multicolumn{1}{c|}{1280} & \multicolumn{1}{c|}{42.76} & \multicolumn{1}{c|}{3.10}        & \multicolumn{1}{c|}{210.67} & \multicolumn{1}{c|}{7.27}        & \multicolumn{1}{c|}{123.67}   & \multicolumn{1}{c|}{6.61}            & \multicolumn{1}{c|}{103.13} & 3.75          \\ \hline
\end{tabular}}
\label{tb:cs1}
\end{table}

\subsubsection{Binary ensemble matrix}

It is shown in~\cite{bar08} that, with high probability, the binary ensemble matrix satisfies the restricted isometry property (RIP)~\cite{ca2008} which guarantees that the sparse signal can be recovered.
Numerical experiments have also shown that binary ensembles performs equally well as Gaussian ensembles in recovering sparse signals~\cite{csbm}.

Each matrix entry is independently chosen to be 1 or $-1$ with probability 1/2.  The rows are then orthonormalized to produce the matrix $\mat{A}$~\cite{nosr}.

We experimented with the same sets of values of $n$, $k$, and $s$ as in the case of Gaussian ensembles.
All solvers are consistently outperformed by their hybrids with our method. The speed up of active-GPSR ranges from 2.16-fold to 74.28-fold, with 
a mean of 25.64-fold. The speedup of active-ADMM  
ranges from 3.51-fold to 367.77-fold, with 
a mean of 83.61-fold. The speedup of active-lassoglm ranges from 3.38-fold to 95.90-fold, with 
a mean of 26.59-fold. The speedup of active-glmnet ranges from 7.61-fold to 79.94-fold, with 
a mean of 29.72-fold.

\begin{table}
\caption{Running time (in seconds) of signal reconstruction using a binary ensemble matrix.  Each running time is an average of 10 trials. }
\resizebox{\columnwidth}{!}{%
\begin{tabular}{|ccccccccccc|}
\hline
\multicolumn{11}{|c|}{Running Time of Signal Reconstruction with Binary Ensemble   Matrix}                                                                                                                                                                                                                                                                                         \\ \hline
\multicolumn{1}{|c|}{n}                         & \multicolumn{1}{c|}{k}                         & \multicolumn{1}{c|}{s}    & \multicolumn{1}{c|}{GPSR}  & \multicolumn{1}{c|}{active-GPSR} & \multicolumn{1}{c|}{ADMM}   & \multicolumn{1}{c|}{active-ADMM} & \multicolumn{1}{c|}{lassoglm} & \multicolumn{1}{c|}{active-lassoglm} & \multicolumn{1}{c|}{glmnet} & active-glmnet \\ \hline
\multicolumn{1}{|c|}{$2^{12}$}                  & \multicolumn{1}{c|}{$2^{10}$}                  & \multicolumn{1}{c|}{160}  & \multicolumn{1}{c|}{0.76}  & \multicolumn{1}{c|}{0.35}        & \multicolumn{1}{c|}{3.19}   & \multicolumn{1}{c|}{0.91}        & \multicolumn{1}{c|}{3.71}     & \multicolumn{1}{c|}{1.10}            & \multicolumn{1}{c|}{2.30}   & 0.30          \\ \hline
\multicolumn{1}{|c|}{\multirow{2}{*}{$2^{13}$}} & \multicolumn{1}{c|}{\multirow{2}{*}{$2^{11}$}} & \multicolumn{1}{c|}{160}  & \multicolumn{1}{c|}{1.45}  & \multicolumn{1}{c|}{0.05}        & \multicolumn{1}{c|}{6.48}   & \multicolumn{1}{c|}{0.15}        & \multicolumn{1}{c|}{4.43}     & \multicolumn{1}{c|}{0.24}            & \multicolumn{1}{c|}{3.95}   & 0.20          \\ \cline{3-11} 
\multicolumn{1}{|c|}{}                          & \multicolumn{1}{c|}{}                          & \multicolumn{1}{c|}{320}  & \multicolumn{1}{c|}{3.08}  & \multicolumn{1}{c|}{0.73}        & \multicolumn{1}{c|}{13.35}  & \multicolumn{1}{c|}{1.60}        & \multicolumn{1}{c|}{11.07}    & \multicolumn{1}{c|}{1.66}            & \multicolumn{1}{c|}{8.28}   & 0.91          \\ \hline
\multicolumn{1}{|c|}{\multirow{4}{*}{$2^{14}$}} & \multicolumn{1}{c|}{\multirow{4}{*}{$2^{12}$}} & \multicolumn{1}{c|}{160}  & \multicolumn{1}{c|}{3.83}  & \multicolumn{1}{c|}{0.08}        & \multicolumn{1}{c|}{17.01}  & \multicolumn{1}{c|}{0.09}        & \multicolumn{1}{c|}{7.46}     & \multicolumn{1}{c|}{0.14}            & \multicolumn{1}{c|}{8.80}   & 0.47          \\ \cline{3-11} 
\multicolumn{1}{|c|}{}                          & \multicolumn{1}{c|}{}                          & \multicolumn{1}{c|}{320}  & \multicolumn{1}{c|}{5.78}  & \multicolumn{1}{c|}{0.19}        & \multicolumn{1}{c|}{26.01}  & \multicolumn{1}{c|}{0.40}        & \multicolumn{1}{c|}{13.49}    & \multicolumn{1}{c|}{0.51}            & \multicolumn{1}{c|}{15.05}  & 0.62          \\ \cline{3-11} 
\multicolumn{1}{|c|}{}                          & \multicolumn{1}{c|}{}                          & \multicolumn{1}{c|}{480}  & \multicolumn{1}{c|}{8.35}  & \multicolumn{1}{c|}{0.56}        & \multicolumn{1}{c|}{36.74}  & \multicolumn{1}{c|}{1.54}        & \multicolumn{1}{c|}{19.48}    & \multicolumn{1}{c|}{1.81}            & \multicolumn{1}{c|}{20.55}  & 1.26          \\ \cline{3-11} 
\multicolumn{1}{|c|}{}                          & \multicolumn{1}{c|}{}                          & \multicolumn{1}{c|}{640}  & \multicolumn{1}{c|}{11.68} & \multicolumn{1}{c|}{2.90}        & \multicolumn{1}{c|}{51.52}  & \multicolumn{1}{c|}{4.86}        & \multicolumn{1}{c|}{29.88}    & \multicolumn{1}{c|}{4.32}            & \multicolumn{1}{c|}{27.71}  & 2.64          \\ \hline
\multicolumn{1}{|c|}{\multirow{8}{*}{$2^{15}$}} & \multicolumn{1}{c|}{\multirow{8}{*}{$2^{13}$}} & \multicolumn{1}{c|}{160}  & \multicolumn{1}{c|}{10.48} & \multicolumn{1}{c|}{0.14}        & \multicolumn{1}{c|}{51.66}  & \multicolumn{1}{c|}{0.14}        & \multicolumn{1}{c|}{17.47}    & \multicolumn{1}{c|}{0.18}            & \multicolumn{1}{c|}{13.69}  & 0.25          \\ \cline{3-11} 
\multicolumn{1}{|c|}{}                          & \multicolumn{1}{c|}{}                          & \multicolumn{1}{c|}{320}  & \multicolumn{1}{c|}{15.13} & \multicolumn{1}{c|}{0.29}        & \multicolumn{1}{c|}{71.19}  & \multicolumn{1}{c|}{0.31}        & \multicolumn{1}{c|}{27.15}    & \multicolumn{1}{c|}{0.47}            & \multicolumn{1}{c|}{33.09}  & 0.41          \\ \cline{3-11} 
\multicolumn{1}{|c|}{}                          & \multicolumn{1}{c|}{}                          & \multicolumn{1}{c|}{480}  & \multicolumn{1}{c|}{18.77} & \multicolumn{1}{c|}{0.48}        & \multicolumn{1}{c|}{87.06}  & \multicolumn{1}{c|}{0.80}        & \multicolumn{1}{c|}{36.45}    & \multicolumn{1}{c|}{1.09}            & \multicolumn{1}{c|}{43.44}  & 0.73          \\ \cline{3-11} 
\multicolumn{1}{|c|}{}                          & \multicolumn{1}{c|}{}                          & \multicolumn{1}{c|}{640}  & \multicolumn{1}{c|}{22.21} & \multicolumn{1}{c|}{0.73}        & \multicolumn{1}{c|}{102.65} & \multicolumn{1}{c|}{1.17}        & \multicolumn{1}{c|}{46.99}    & \multicolumn{1}{c|}{1.43}            & \multicolumn{1}{c|}{52.31}  & 0.97          \\ \cline{3-11} 
\multicolumn{1}{|c|}{}                          & \multicolumn{1}{c|}{}                          & \multicolumn{1}{c|}{800}  & \multicolumn{1}{c|}{26.79} & \multicolumn{1}{c|}{1.28}        & \multicolumn{1}{c|}{124.09} & \multicolumn{1}{c|}{2.53}        & \multicolumn{1}{c|}{57.92}    & \multicolumn{1}{c|}{3.44}            & \multicolumn{1}{c|}{62.50}  & 1.97          \\ \cline{3-11} 
\multicolumn{1}{|c|}{}                          & \multicolumn{1}{c|}{}                          & \multicolumn{1}{c|}{960}  & \multicolumn{1}{c|}{31.76} & \multicolumn{1}{c|}{1.91}        & \multicolumn{1}{c|}{147.64} & \multicolumn{1}{c|}{4.04}        & \multicolumn{1}{c|}{69.07}    & \multicolumn{1}{c|}{4.18}            & \multicolumn{1}{c|}{79.54}  & 2.75          \\ \cline{3-11} 
\multicolumn{1}{|c|}{}                          & \multicolumn{1}{c|}{}                          & \multicolumn{1}{c|}{1120} & \multicolumn{1}{c|}{38.38} & \multicolumn{1}{c|}{4.51}        & \multicolumn{1}{c|}{175.76} & \multicolumn{1}{c|}{9.08}        & \multicolumn{1}{c|}{88.72}    & \multicolumn{1}{c|}{7.77}            & \multicolumn{1}{c|}{96.15}  & 5.18          \\ \cline{3-11} 
\multicolumn{1}{|c|}{}                          & \multicolumn{1}{c|}{}                          & \multicolumn{1}{c|}{1280} & \multicolumn{1}{c|}{46.10} & \multicolumn{1}{c|}{11.12}       & \multicolumn{1}{c|}{207.62} & \multicolumn{1}{c|}{16.07}       & \multicolumn{1}{c|}{104.49}   & \multicolumn{1}{c|}{13.16}           & \multicolumn{1}{c|}{111.61} & 9.04          \\ \hline
\end{tabular}}
\label{tb:cs2}
\end{table}

\subsection{Lasso Regression}

The Lasso regression is to solve $ \frac{1}{2}\|\mat A\mat x - \mat b \|^2_2 + \eta \|\mat x \|_1$ for some given $\mat{A}$ and $\mat{b}$.  We experimented with several settings.  Although the least square term has a coefficient of $1/2$ instead of 1, our algorithm is still applicable.

\paragraph{Regression with Noise.}

In the absence of ground truth, a dataset is often split into two halves, the \emph{training set} and the \emph{validation set}, for selecting an appropriate value for $\beta$.  The training set induces a matrix $\mat{A}_1$ and vector $\mat{b}_1$.  The validation set induces another matrix $\mat{A}_2$ and vector $\mat{b}_2$.  These matrices and vectors have the same row dimensions as the training and validation sets have the same size.

There is a list of candidate values for $\eta$.  For each candidate value, we solve $\frac{1}{2}\|\mat{A}_1\mat x - \mat{b}_1 \|^2_2 + \eta \|\mat x \|_1$ for $\mat{x}$, and then compute the mean square error $\frac{1}{n}\|\mat{A}_2\mat{x} - \mat{b}_2\|_2^2$, where $n$ is the size of the validation set.  The desired value for $\eta$ is the one that yields the smallest mean square error.  We study the efficiency of active-GPSR, active-lassoglm, and active-glmnet on this task.

We use an experimental setup akin to the one in~\cite{ADMM_re}. The matrix $\mat A$ is $n \times d$, where $n=6,000$ and $d=120,000$.  Each entry of $\mat{A}$ is drawn uniformly at random from $[0,1]$.  The sparse ground truth $\mat{z} \in \mathbb{R}^d$ is produced by choosing 150 coordinates independently and uniformly at random, and drawing each of these 150 coordinates uniformly at random from $[0,1]$. Then we set $\mat{b} = \mat A {\mat z} + \mat n$, where $\mat n$ is the Gaussian white noise, and each entry of $\mat{n}$ has mean 0 and variance $0.1$.

In all runs, $\eta = 18$ yields the smallest mean square error for all solvers and their hybrids with our method. 
 Table~\ref{tb:genlin} shows the running times for the candidate values of $\eta$ tested.  The speedup of active-GPSR ranges from 15.29
-fold to 27.91-fold, with 
a mean of 21.16-fold. The speedup of active-lassoglm 
ranges from 2.48-fold to 11.99-fold, with 
a mean of 8.06-fold. The speedup of active-glmnet ranges from 15.20-fold to 25.50-fold, with 
a mean of 19.55-fold.

\begin{table}
\centering
\caption{Running time (in seconds) of parameter selection for Lasso regression on synthetic data with noise.  The data is randomly generated.  Each running time is an average of 10 trials.}
 \resizebox{.85\columnwidth}{!}{%
\begin{tabular}{|ccccccc|}
\hline
\multicolumn{7}{|c|}{Parameter Selection for Lasso Regression on Synthetic Data with Noise}                                                                                                                                                        \\ \hline
\multicolumn{1}{|c|}{$\eta$} & \multicolumn{1}{c|}{GPSR}   & \multicolumn{1}{c|}{active-GPSR} & \multicolumn{1}{c|}{lassoglm} & \multicolumn{1}{c|}{active-lassoglm} & \multicolumn{1}{c|}{glmnet} & active-glmnet \\ \hline
\multicolumn{1}{|c|}{9}   & \multicolumn{1}{c|}{189.20} & \multicolumn{1}{c|}{12.38}       & \multicolumn{1}{c|}{102.33}   & \multicolumn{1}{c|}{37.95}           & \multicolumn{1}{c|}{132.14} & 8.69          \\ \hline
\multicolumn{1}{|c|}{10}  & \multicolumn{1}{c|}{161.60} & \multicolumn{1}{c|}{10.16}       & \multicolumn{1}{c|}{73.38}    & \multicolumn{1}{c|}{29.59}           & \multicolumn{1}{c|}{117.82} & 6.68          \\ \hline
\multicolumn{1}{|c|}{11}  & \multicolumn{1}{c|}{139.26} & \multicolumn{1}{c|}{8.61}        & \multicolumn{1}{c|}{70.03}    & \multicolumn{1}{c|}{21.09}           & \multicolumn{1}{c|}{106.30} & 6.14          \\ \hline
\multicolumn{1}{|c|}{12}  & \multicolumn{1}{c|}{119.36} & \multicolumn{1}{c|}{7.22}        & \multicolumn{1}{c|}{69.77}    & \multicolumn{1}{c|}{20.40}           & \multicolumn{1}{c|}{95.14}  & 5.28          \\ \hline
\multicolumn{1}{|c|}{13}  & \multicolumn{1}{c|}{105.85} & \multicolumn{1}{c|}{5.81}        & \multicolumn{1}{c|}{188.37}   & \multicolumn{1}{c|}{24.05}           & \multicolumn{1}{c|}{85.32}  & 5.04          \\ \hline
\multicolumn{1}{|c|}{14}  & \multicolumn{1}{c|}{86.40}  & \multicolumn{1}{c|}{5.07}        & \multicolumn{1}{c|}{149.77}   & \multicolumn{1}{c|}{17.70}           & \multicolumn{1}{c|}{77.94}  & 3.76          \\ \hline
\multicolumn{1}{|c|}{15}  & \multicolumn{1}{c|}{83.10}  & \multicolumn{1}{c|}{3.71}        & \multicolumn{1}{c|}{136.74}   & \multicolumn{1}{c|}{16.93}           & \multicolumn{1}{c|}{66.46}  & 3.50          \\ \hline
\multicolumn{1}{|c|}{16}  & \multicolumn{1}{c|}{69.39}  & \multicolumn{1}{c|}{3.41}        & \multicolumn{1}{c|}{137.43}   & \multicolumn{1}{c|}{14.20}           & \multicolumn{1}{c|}{65.71}  & 2.81          \\ \hline
\multicolumn{1}{|c|}{17}  & \multicolumn{1}{c|}{65.33}  & \multicolumn{1}{c|}{2.61}        & \multicolumn{1}{c|}{116.76}   & \multicolumn{1}{c|}{11.52}           & \multicolumn{1}{c|}{61.62}  & 3.18          \\ \hline
\multicolumn{1}{|c|}{18}  & \multicolumn{1}{c|}{61.15}  & \multicolumn{1}{c|}{2.40}        & \multicolumn{1}{c|}{102.65}   & \multicolumn{1}{c|}{10.69}           & \multicolumn{1}{c|}{57.79}  & 2.60          \\ \hline
\multicolumn{1}{|c|}{19}  & \multicolumn{1}{c|}{52.64}  & \multicolumn{1}{c|}{1.89}        & \multicolumn{1}{c|}{96.12}    & \multicolumn{1}{c|}{10.37}           & \multicolumn{1}{c|}{53.20}  & 2.13          \\ \hline
\multicolumn{1}{|c|}{20}  & \multicolumn{1}{c|}{42.49}  & \multicolumn{1}{c|}{1.72}        & \multicolumn{1}{c|}{91.97}    & \multicolumn{1}{c|}{7.67}            & \multicolumn{1}{c|}{51.94}  & 2.04          \\ \hline
\multicolumn{1}{|c|}{21}  & \multicolumn{1}{c|}{42.35}  & \multicolumn{1}{c|}{2.06}        & \multicolumn{1}{c|}{82.74}    & \multicolumn{1}{c|}{9.53}            & \multicolumn{1}{c|}{47.11}  & 2.47          \\ \hline
\multicolumn{1}{|c|}{22}  & \multicolumn{1}{c|}{41.62}  & \multicolumn{1}{c|}{1.90}        & \multicolumn{1}{c|}{78.29}    & \multicolumn{1}{c|}{9.01}            & \multicolumn{1}{c|}{42.55}  & 2.35          \\ \hline
\multicolumn{1}{|c|}{23}  & \multicolumn{1}{c|}{37.41}  & \multicolumn{1}{c|}{1.81}        & \multicolumn{1}{c|}{77.23}    & \multicolumn{1}{c|}{8.75}            & \multicolumn{1}{c|}{39.17}  & 2.34          \\ \hline
\multicolumn{1}{|c|}{24}  & \multicolumn{1}{c|}{34.91}  & \multicolumn{1}{c|}{1.78}        & \multicolumn{1}{c|}{73.47}    & \multicolumn{1}{c|}{8.11}            & \multicolumn{1}{c|}{38.03}  & 1.91          \\ \hline
\multicolumn{1}{|c|}{25}  & \multicolumn{1}{c|}{35.86}  & \multicolumn{1}{c|}{1.39}        & \multicolumn{1}{c|}{66.87}    & \multicolumn{1}{c|}{7.39}            & \multicolumn{1}{c|}{36.65}  & 1.85          \\ \hline
\multicolumn{1}{|c|}{26}  & \multicolumn{1}{c|}{30.97}  & \multicolumn{1}{c|}{1.35}        & \multicolumn{1}{c|}{67.67}    & \multicolumn{1}{c|}{6.89}            & \multicolumn{1}{c|}{33.18}  & 1.79          \\ \hline
\multicolumn{1}{|c|}{27}  & \multicolumn{1}{c|}{26.24}  & \multicolumn{1}{c|}{1.30}        & \multicolumn{1}{c|}{66.39}    & \multicolumn{1}{c|}{6.74}            & \multicolumn{1}{c|}{33.51}  & 1.70          \\ \hline
\multicolumn{1}{|c|}{28}  & \multicolumn{1}{c|}{33.36}  & \multicolumn{1}{c|}{1.27}        & \multicolumn{1}{c|}{64.85}    & \multicolumn{1}{c|}{6.39}            & \multicolumn{1}{c|}{31.37}  & 1.66          \\ \hline
\end{tabular}}
	\label{tb:genlin}
\end{table}

\paragraph{The E2006 dataset.}

We also tested parameter selection using the real high-dimensional data set
E2006-tfidf~\cite{libsvm, sdb09}.  It contains $2n = 16,087$ samples, and each sample has $d = 150,360$ features. As before, we split the dataset into two halves, the training set and the validation set, which give rise to matrix $\mat{A}_1 \in \mathbb{R}^{n \times d}$ and vector $\mat{b}_1 \in \mathbb{R}^d$ for the training set, and matrix $\mat{A}_2 \in \mathbb{R}^{n \times d}$ and vector $\mat{b}_2 \in \mathbb{R}^d$ for the validation set.

Following \cite{lasso}, we standardize each column of $\mat{A}_1$. That is, for the $j$-th column, we compute the mean $\alpha_{1,j}$ and standard deviation $\sigma_{1,j}$ of its entries, and replace each entry $(\mat{A}_1)_{i,j}$ in the $j$-th column by $((\mat{A}_1)_{i,j} - \alpha_{1,j})/\sigma_{1,j}$.  Similarly, each entry $(\mat{b}_1)_i$ of $\mat{b}_1$ is replaced by $(\mat{b}_1)_i - \beta_1$, where $\beta_1$ is the mean of the entries of $\mat{b}_1$, respectively.
This is a routine step for removing the measurement scale effect.  Correspondingly, each entry $(\mat{A}_2)_{i,j}$ of the $j$-th column of $\mat{A}_2$ is replaced by $((\mat{A}_2)_{i,j} - \alpha_{1,j})/\sigma_{1,j}$, and each entry $(\mat{b}_2)_i$ of $\mat{b}_2$ is replaced by $(\mat{b}_2)_i - \beta_1$.  Afterwards, we select the best value for $\eta$ as explained before.

Table~\ref{tb:lr1} shows the running times for the candidate values of $\eta$ tested.  The speedup of active-GPSR ranges from 21.24-fold to 84.21-fold, with 
a mean of 54.87-fold. The speedup of active-lassoglm ranges from 21.51-fold to 75.67-fold, with 
a mean of 40.05-fold. The speedup of active-glmnet ranges from 2.84-fold to 8.44-fold, with 
a mean of 6.30-fold. 

\begin{table}
\centering
\caption{Running time (in seconds) of parameter selection for Lasso Regression on E2006-tfidf.  Each running time is an average of 3 trials.}
\resizebox{.85\columnwidth}{!}{%
\begin{tabular}{|ccccccc|}
\hline
\multicolumn{7}{|c|}{Parameter   selection for Lasso Regression on E2006-tfidf}                                                                                                                                     \\ \hline
\multicolumn{1}{|c|}{$\eta$} & \multicolumn{1}{c|}{GPSR}    & \multicolumn{1}{c|}{active-GPSR} & \multicolumn{1}{c|}{lassoglm} & \multicolumn{1}{c|}{active-lassoglm} & \multicolumn{1}{c|}{glmnet} & active-glmnet \\ \hline
\multicolumn{1}{|c|}{300}    & \multicolumn{1}{c|}{92.89}   & \multicolumn{1}{c|}{1.88}        & \multicolumn{1}{c|}{54.35}    & \multicolumn{1}{c|}{2.34}            & \multicolumn{1}{c|}{10.28}  & 1.22          \\ \hline
\multicolumn{1}{|c|}{280}    & \multicolumn{1}{c|}{139.93}  & \multicolumn{1}{c|}{2.25}        & \multicolumn{1}{c|}{60.38}    & \multicolumn{1}{c|}{2.45}            & \multicolumn{1}{c|}{10.30}  & 1.23          \\ \hline
\multicolumn{1}{|c|}{260}    & \multicolumn{1}{c|}{150.84}  & \multicolumn{1}{c|}{2.76}        & \multicolumn{1}{c|}{84.71}    & \multicolumn{1}{c|}{2.61}            & \multicolumn{1}{c|}{10.90}  & 1.29          \\ \hline
\multicolumn{1}{|c|}{240}    & \multicolumn{1}{c|}{665.94}  & \multicolumn{1}{c|}{10.93}       & \multicolumn{1}{c|}{76.87}    & \multicolumn{1}{c|}{2.71}            & \multicolumn{1}{c|}{10.49}  & 1.33          \\ \hline
\multicolumn{1}{|c|}{220}    & \multicolumn{1}{c|}{2091.35} & \multicolumn{1}{c|}{27.49}       & \multicolumn{1}{c|}{83.64}    & \multicolumn{1}{c|}{2.98}            & \multicolumn{1}{c|}{10.48}  & 1.38          \\ \hline
\multicolumn{1}{|c|}{200}    & \multicolumn{1}{c|}{2433.30} & \multicolumn{1}{c|}{28.90}       & \multicolumn{1}{c|}{353.14}   & \multicolumn{1}{c|}{5.05}            & \multicolumn{1}{c|}{9.38}   & 1.40          \\ \hline
\multicolumn{1}{|c|}{180}    & \multicolumn{1}{c|}{2840.02} & \multicolumn{1}{c|}{37.37}       & \multicolumn{1}{c|}{380.69}   & \multicolumn{1}{c|}{5.52}            & \multicolumn{1}{c|}{9.55}   & 1.53          \\ \hline
\multicolumn{1}{|c|}{160}    & \multicolumn{1}{c|}{3255.01} & \multicolumn{1}{c|}{59.78}       & \multicolumn{1}{c|}{658.77}   & \multicolumn{1}{c|}{8.71}            & \multicolumn{1}{c|}{9.77}   & 1.78          \\ \hline
\multicolumn{1}{|c|}{140}    & \multicolumn{1}{c|}{5704.94} & \multicolumn{1}{c|}{149.08}      & \multicolumn{1}{c|}{652.82}   & \multicolumn{1}{c|}{15.80}           & \multicolumn{1}{c|}{9.57}   & 2.24          \\ \hline
\multicolumn{1}{|c|}{120}    & \multicolumn{1}{c|}{5703.87} & \multicolumn{1}{c|}{218.42}      & \multicolumn{1}{c|}{665.96}   & \multicolumn{1}{c|}{25.09}           & \multicolumn{1}{c|}{10.99}  & 3.59          \\ \hline
\multicolumn{1}{|c|}{100}    & \multicolumn{1}{c|}{5712.09} & \multicolumn{1}{c|}{268.96}      & \multicolumn{1}{c|}{690.07}   & \multicolumn{1}{c|}{32.09}           & \multicolumn{1}{c|}{11.88}  & 4.19          \\ \hline
\end{tabular}}
	\label{tb:lr1}
\end{table}

\paragraph{Regression with Correlated Variables.} We also experimented with synthetic data that consists of correlated explanatory variables.  As described in~\cite{glmlassopaper}, one can generate a matrix $\mat A \in \mathbb{R}^{n \times d}$ with entries sampled from a Gaussian distribution such that the population correlation between any two explanatory variables is $\rho$, and define the vector $\mat b = \mat{A}\mat{z}+ k\mat{n}$, where each entry $(\mat z)_j$ is equal to $(-1)^j \exp(-2(j-1)/20)$, each entry of $\mat{n}$ is sampled independently from the standard normal distribution, and $k$ is chosen to make the signal-to-noise ratio equal to $3.0$. We used the procedure available on the official page of the paper~\cite{glmlassopaper} to generate such a dataset.

For each configuration of $(n, d, \rho)$, we select the best value for $\eta$ by calling a built-in function of lassoglm that performs a 5-fold cross-validation on the dataset.
After fixing $\eta$, we run the solvers and their hybrids with our method one last time on the complete dataset. 
 Table~\ref{tb:lassoCorr} shows the running times of the last runs on the complete dataset.  The speedup of active-GPSR ranges from 5.10-fold to 30.29-fold, with 
a mean of 15.99-fold. The speedup of active-lassoglm ranges from 1.06-fold to 3.16-fold, with 
a mean of 1.89-fold. The speedup of active-glmnet ranges from 1.60-fold to 5.81-fold, with 
a mean of 3.11-fold.

\begin{table}
\caption{Running time (in seconds) of Lasso Regression on synthetic data with correlated explanatory variables.  Each running time is an average of 10 trials.}
\centering
 \resizebox{.85\columnwidth}{!}{%
\begin{tabular}{|ccccccccc|}
\hline
\multicolumn{9}{|c|}{Lasso Regression on Synthetic Data with Correlated Explanatory Variables}                                                                                                                                                                                                                                                                \\ \hline
\multicolumn{1}{|l|}{n}                     & \multicolumn{1}{l|}{d}                      & \multicolumn{1}{l|}{$\rho$} & \multicolumn{1}{l|}{GPSR}   & \multicolumn{1}{l|}{active-GPSR} & \multicolumn{1}{l|}{lassoglm} & \multicolumn{1}{l|}{active-lassoglm} & \multicolumn{1}{l|}{glmnet} & \multicolumn{1}{l|}{active-glmnet} \\ \hline
\multicolumn{1}{|c|}{\multirow{6}{*}{1250}} & \multicolumn{1}{c|}{\multirow{3}{*}{7500}}  & \multicolumn{1}{c|}{0.3} & \multicolumn{1}{c|}{9.33}   & \multicolumn{1}{c|}{0.36}        & \multicolumn{1}{c|}{6.17}     & \multicolumn{1}{c|}{2.93}            & \multicolumn{1}{c|}{1.45}   & 0.57                               \\ \cline{3-9} 
\multicolumn{1}{|c|}{}                      & \multicolumn{1}{c|}{}                       & \multicolumn{1}{c|}{0.6} & \multicolumn{1}{c|}{24.22}  & \multicolumn{1}{c|}{0.98}        & \multicolumn{1}{c|}{11.04}    & \multicolumn{1}{c|}{8.36}            & \multicolumn{1}{c|}{22.65}  & 6.93                               \\ \cline{3-9} 
\multicolumn{1}{|c|}{}                      & \multicolumn{1}{c|}{}                       & \multicolumn{1}{c|}{0.9} & \multicolumn{1}{c|}{42.26}  & \multicolumn{1}{c|}{2.75}        & \multicolumn{1}{c|}{10.96}    & \multicolumn{1}{c|}{10.36}           & \multicolumn{1}{c|}{25.73}  & 4.64                               \\ \cline{2-9} 
\multicolumn{1}{|c|}{}                      & \multicolumn{1}{c|}{\multirow{3}{*}{10000}} & \multicolumn{1}{c|}{0.3} & \multicolumn{1}{c|}{22.46}  & \multicolumn{1}{c|}{0.74}        & \multicolumn{1}{c|}{9.13}     & \multicolumn{1}{c|}{4.92}            & \multicolumn{1}{c|}{3.80}   & 2.02                               \\ \cline{3-9} 
\multicolumn{1}{|c|}{}                      & \multicolumn{1}{c|}{}                       & \multicolumn{1}{c|}{0.6} & \multicolumn{1}{c|}{57.47}  & \multicolumn{1}{c|}{2.76}        & \multicolumn{1}{c|}{16.70}    & \multicolumn{1}{c|}{12.36}           & \multicolumn{1}{c|}{39.57}  & 9.98                               \\ \cline{3-9} 
\multicolumn{1}{|c|}{}                      & \multicolumn{1}{c|}{}                       & \multicolumn{1}{c|}{0.9} & \multicolumn{1}{c|}{57.90}  & \multicolumn{1}{c|}{3.47}        & \multicolumn{1}{c|}{19.31}    & \multicolumn{1}{c|}{13.17}           & \multicolumn{1}{c|}{84.54}  & 14.55                              \\ \hline
\multicolumn{1}{|c|}{\multirow{6}{*}{2500}} & \multicolumn{1}{c|}{\multirow{3}{*}{7500}}  & \multicolumn{1}{c|}{0.3} & \multicolumn{1}{c|}{16.49}  & \multicolumn{1}{c|}{1.32}        & \multicolumn{1}{c|}{12.23}    & \multicolumn{1}{c|}{4.28}            & \multicolumn{1}{c|}{3.93}   & 2.38                               \\ \cline{3-9} 
\multicolumn{1}{|c|}{}                      & \multicolumn{1}{c|}{}                       & \multicolumn{1}{c|}{0.6} & \multicolumn{1}{c|}{54.45}  & \multicolumn{1}{c|}{5.19}        & \multicolumn{1}{c|}{13.62}    & \multicolumn{1}{c|}{5.40}            & \multicolumn{1}{c|}{72.15}  & 21.34                              \\ \cline{3-9} 
\multicolumn{1}{|c|}{}                      & \multicolumn{1}{c|}{}                       & \multicolumn{1}{c|}{0.9} & \multicolumn{1}{c|}{85.24}  & \multicolumn{1}{c|}{11.20}       & \multicolumn{1}{c|}{14.01}    & \multicolumn{1}{c|}{6.99}            & \multicolumn{1}{c|}{15.84}  & 8.80                               \\ \cline{2-9} 
\multicolumn{1}{|c|}{}                      & \multicolumn{1}{c|}{\multirow{3}{*}{10000}} & \multicolumn{1}{c|}{0.3} & \multicolumn{1}{c|}{30.14}  & \multicolumn{1}{c|}{2.61}        & \multicolumn{1}{c|}{17.18}    & \multicolumn{1}{c|}{5.44}            & \multicolumn{1}{c|}{4.93}   & 3.08                               \\ \cline{3-9} 
\multicolumn{1}{|c|}{}                      & \multicolumn{1}{c|}{}                       & \multicolumn{1}{c|}{0.6} & \multicolumn{1}{c|}{108.44} & \multicolumn{1}{c|}{9.68}        & \multicolumn{1}{c|}{19.14}    & \multicolumn{1}{c|}{10.91}           & \multicolumn{1}{c|}{56.27}  & 19.35                              \\ \cline{3-9} 
\multicolumn{1}{|c|}{}                      & \multicolumn{1}{c|}{}                       & \multicolumn{1}{c|}{0.9} & \multicolumn{1}{c|}{114.80} & \multicolumn{1}{c|}{22.53}       & \multicolumn{1}{c|}{18.39}    & \multicolumn{1}{c|}{15.47}           & \multicolumn{1}{c|}{40.03}  & 13.75                              \\ \hline
\end{tabular}}
	\label{tb:lassoCorr}
\end{table}

\subsection{Logistic Lasso Regression}

Recall that Logistic Lasso regression is to solve $\min -\frac{1}{n}\sum^n_{i=1}  \bigl(y_i \log \mu_i + (1-y_i) \log(1-\mu_i)\bigr) +\eta \|\mat x\|_1$, where $\mu_i = 1/(1+\mbox{exp}(-\mat{a}_i^t \mat{x}))$.

We used three datasets from the LIBSVM library~\cite{libsvm}: url\_combined\_normalized~\cite{jm09a}, criteo~\cite{criteo}, and kdda~\cite{kddacite}. Given the large sizes of these datasets and the extensive memory requirements of lassoglm, for each dataset, we randomly picked a subset for a 5-fold cross-validation to select the best value for $\eta$.
For url\_combined\_normalized and criteo, we randomly picked $n=5,000$ samples with $d=20,000$ features.  For kdda, we randomly picked $n=10,000$ samples with $ d=40,000$ features.  We called a built-in function of lassoglm for the 5-fold cross-validation.

After fixing $\eta$, we compare the efficiency of lassoglm, active-lassoglm, glmnet, and active-glmnet in the final training.  As noted in~\cite{mlpp}, final training is a common technique in machine learning, which entails solving for $\mat{x}$ on larger data.
For the final training, we draw $n=10,000$ samples from both url\_combined\_normalized and criteo, and we draw $n=20,000$ samples from kdda.  Table~\ref{tb:log1} shows the running times of the final training.
The speedup of active-lassoglm ranges from 9.72-fold to 13.58-fold, with 
a mean of 12.53-fold. The speedup of active-glmnet ranges from 1.23-fold to 1.56-fold, with 
a mean of 1.41-fold.

\begin{table}
\centering
\caption{Final training time (in seconds) for Logistic Lasso Regression. Each running time is an average of 10 trials. }
\resizebox{.85\columnwidth}{!}{%
\begin{tabular}{|ccccc|}
\hline
\multicolumn{5}{|c|}{Final Training Time for Logistic Lasso Regression}                                                                          \\ \hline
\multicolumn{1}{|c|}{}                          & \multicolumn{1}{c|}{lassoglm} & \multicolumn{1}{c|}{active-lassoglm} & \multicolumn{1}{c|}{glmnet} & active-glmnet \\ \hline
\multicolumn{1}{|l|}{url   combined normalized} & \multicolumn{1}{c|}{9.83}     & \multicolumn{1}{c|}{1.01}            & \multicolumn{1}{c|}{0.86}   & 0.70          \\ \hline
\multicolumn{1}{|l|}{criteo}                    & \multicolumn{1}{c|}{10.35}    & \multicolumn{1}{c|}{0.83}            & \multicolumn{1}{c|}{0.85}   & 0.60          \\ \hline
\multicolumn{1}{|l|}{kdda}                      & \multicolumn{1}{c|}{68.54}    & \multicolumn{1}{c|}{5.05}            & \multicolumn{1}{c|}{4.04}   & 2.58          \\ \hline
\end{tabular}}
\label{tb:log1}
\end{table}

\section{Conclusion}

The hybrids of our method with GPSR, lassoglm, ADMM, and glmnet show significant speedups.
Our rule for updating the active set between iterations has well-founded theoretical underpinnings.  In most cases, the rule frees a tiny number of variables in the active set between iterations, which makes each call of the solver fast. There is a lot of room to enhance efficiency because we plainly reinitialize before each solver call.  A more careful integration should yield a higher efficiency.  Another question is to mathematically analyze the convergence rate of our algorithm.

\section{Impact Statements}
This paper presents work whose goal is to advance the field of Machine Learning. There are many potential societal consequences of our work, none which we feel must be specifically highlighted here.

\bibliographystyle{plain}
\bibliography{main}


\appendix

\section{URLs to the Solver Pages}
Here is a list of URLs to the solver pages:
\begin{itemize}
    \item GPSR: \href{http://www.lx.it.pt/~mtf/GPSR/}{http://www.lx.it.pt/$\sim$mtf/GPSRR/}
    \item ADMM: \href{https://web.stanford.edu/~boyd/papers/admm/lasso/lasso.html}{https://web.stanford.edu/$\sim$boyd/papers/admm/lasso/lasso.html}
    \item lassoglm: \href{https://www.mathworks.com/help/stats/lassoglm.html}{https://www.mathworks.com/help/stats/lassoglm.html}
    \item glmnet: \href{https://hastie.su.domains/glmnet_matlab/download.html}{https://hastie.su.domains/glmnet\_matlab/download.html}
    \item For glmnet with compilation on a later version of Windows, used in the experiments:
    
    \href{https://github.com/lachioma/glmnet_matlab}{https://github.com/lachioma/glmnet\_matlab}
\end{itemize}
Note that these links are provided only for courtesy purposes. The authors do not have any direct or indirect control over the public pages and are unaffiliated.

\section{Experiments}

\begin{table}[h]
\caption{Parameter settings on the base solvers.}
\resizebox{\columnwidth}{!}{
\begin{tabular}{|c|c|c|c|c|}
\hline
\multicolumn{1}{|l|}{} & Default setting          & Compressed Sensing & Lasso Linear Regression & Lasso logistic Regression \\ \hline
GPSR                   & TolA=1E-2                & 1E-05               & 1E-05                   & N/A                       \\ \hline
ADMM                   & ABSTOL=1E-4; RELTOL=1E-4 & Default             & N/A                     & N/A                       \\ \hline
lassoglm               & RelTol=1E-4              & Default             & Default                 & Default                   \\ \hline
glmnet                 & thresh=1E-7              & 1E-10               & 1E-10                & Default                   \\ \hline
\end{tabular}}
\label{tb:parameter-default}
\end{table}

\begin{table}[h]
\caption{Parameter values in our experiments.}
\resizebox{\columnwidth}{!}{
\label{tb:parameters}
\centering
\begin{tabular}{|lcccccc|}
\hline                                                                      
\multicolumn{1}{|l|}{}                & \multicolumn{2}{c|}{Compressed Sensing}                             & \multicolumn{2}{c|}{Lasso Linear Regression}                         & \multicolumn{2}{c|}{Lasso Logistic Regression}  \\ \hline
\multicolumn{1}{|l|}{}                & \multicolumn{1}{p{0.75in}|}{Before extra iteration} & \multicolumn{1}{c|}{Extra iteration} & \multicolumn{1}{p{0.75in}|}{Before extra iteration} & \multicolumn{1}{c|}{Extra iteration}  & \multicolumn{1}{p{0.75in}|}{Before extra iteration} & \multicolumn{1}{c|}{Extra iteration}  \\ \hline
\multicolumn{1}{|l|}{active-GPSRm (TolA)}     & \multicolumn{1}{c|}{1E-03}      & \multicolumn{1}{c|}{1E-05}   & \multicolumn{1}{c|}{1E-03}      & \multicolumn{1}{c|}{1E-05}   & \multicolumn{1}{c|}{N/A}           & N/A        \\ \hline
\multicolumn{1}{|l|}{active-ADMM}     & \multicolumn{1}{c|}{default}       & \multicolumn{1}{c|}{N/A}    & \multicolumn{1}{c|}{N/A}           & \multicolumn{1}{c|}{N/A}        & \multicolumn{1}{c|}{N/A}           & N/A        \\ \hline
\multicolumn{1}{|l|}{active-lassoglm} & \multicolumn{1}{c|}{default}       & \multicolumn{1}{c|}{N/A}    & \multicolumn{1}{c|}{default}       & \multicolumn{1}{c|}{N/A}    & \multicolumn{1}{c|}{default}       & N/A    \\ \hline
\multicolumn{1}{|l|}{active-glmnet (thresh)}   & \multicolumn{1}{c|}{1E-12}      & \multicolumn{1}{c|}{N/A}        & \multicolumn{1}{c|}{1E-05}      & \multicolumn{1}{c|}{1E-10}   & \multicolumn{1}{c|}{default}       & N/A   \\ \hline
\end{tabular}}
\label{tb:subsetsetting}
\end{table}

\section{Angle boosting}

We are given vectors $\mat{w}_i = \varphi(\mat{e}_i)$ for $i \in [\rho]$ such that $\mat{h} = \varphi(\mat{d})$ makes an angle no more than $\arccos(1/\log m)$ with $\mat{w}_i$ for $i \in [\rho]$.  We want to prove that there is a conical combination of $\{\mat{w}_i : i \in [\rho]\}$ that makes an angle no more than $\arccos(4/9$ with $\mat{h}$.  Our proof is an adaption of the proof of an analogous result in~\cite{nnqp}[Lemma~5].  
A key property that we exploit is that, by the dimension reduction property, the volume of any simplex formed at most $2 + \log m$ points in $X$ is preserved under $\varphi$ with distortion $\leq 1+\varepsilon$.  The following technical lemma is a result that we will apply several times.

\vspace{6pt}

\begin{lemma}
\label{lem:tech}
  Let $P$ be a set of vectors $\mat{p}_i$'s of size $\log m$ such that:
  \begin{itemize}
      \item Each $\mat{p}_i$ makes an angle no more than $\arccos(1/N)$ with $\mat{h}$.  
      
      \item The volume of the simplex spanned by 0 and any subset of $P$ is at least $1/(1+\varepsilon)$.
      
  \end{itemize}
  There is a conical combination of $P$ that makes an angle no more than $\arccos(\sqrt{2/3} \cdot (\log m)^{\log_3 5 -1}/N)$ with $\mat{h}$.
\end{lemma}
\begin{proof}
  Let $Q = \{\mat{q}_1,\mat{q}_2,\ldots\}$ be a maximal subset of $P$ with size equal to a power of 2.  Let $\mat{q}_{1,2} = \frac{1}{\sqrt{2}}\mat{q}_1 + \frac{1}{\sqrt{2}}\mat{q}_2$.  Refer to Figure~\ref{fg:dim} which is adapted from an analogous figure in~\cite{nnqp}.  By assumption, the angle between $\mat{h}$ and $\mat{q}_1$ is at most $\theta = \arccos(1/N)$.  Let $\phi$ be the angle between the vectors $\mat{h}$ and $\mat{q}_{1,2}$.  The vector $\mat{q}_{1,2}$ bisects the angle between $\mat{q}_1$ and $\mat{q}_2$.  By assumption, $\sin(2\psi) \geq 1/(1+\varepsilon)$, which implies that $\cos(2\psi) \leq \sqrt{\varepsilon(2+\varepsilon)}/(1+\varepsilon)$.  Hence, $\cos^2\psi \leq \frac{1}{2} + \frac{1}{2} \cdot \sqrt{\varepsilon(2+\varepsilon)}/(1+\varepsilon) \leq 2/3$ for a small enough $\varepsilon$.  The plane spanned by the vectors $\mat{h}$ and $\mat{q}_{1,2}$ splits the angle $\pi$ at $\mat{q}_{1,2}$ into two parts.  Let $\zeta$ be the non-acute part.  Then, by the law of cosines, $\cos \theta \leq \cos \angle (\mat{h},\mat{q}_1) = \cos\phi \cos\psi + \sin\phi \sin(\psi)\cos\zeta \leq \cos\phi\cos\psi$ as $\cos\zeta \leq 0$.  Therefore, $\cos\phi \geq \sqrt{3/2}\cos\theta$. 
 Similarly, the cosine of the angle between $\mat{h}$ and $\mat{q}_{3,4} = \frac{1}{\sqrt{2}}\mat{q}_3 + \frac{1}{\sqrt{2}}\mat{q}_4$ is also at least $\sqrt{3/2}\cos\theta$.  In all, for $i = 1, 2, ..., |Q|/2$, we obtain a unit vector $\mat{q}_{2i-1,2i}$ that makes an angle no more than $\arccos(\sqrt{3/2}\cos\theta)$ with $\hat{h}$.

  \begin{figure}
      \centerline{\includegraphics[scale=0.6]{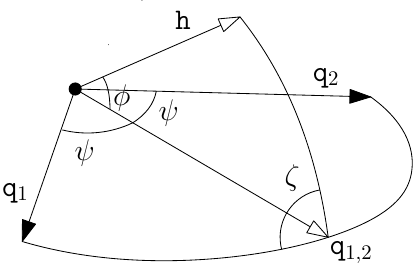}}
      \caption{}
      \label{fg:dim}
  \end{figure}

  Let $\sigma$ be the simplex spanned by $0,\mat{q}_1,\mat{q}_3,\mat{q}_4,\ldots$.  By volume preservation, the volume of $\sigma$ is at least a fraction $1+\varepsilon$ of the volume of the simplex spanned by $0,\mat{e}_1,\mat{e}_3,\mat{e}_4,\ldots$.  The volume of the simplex $\sigma'$ spanned by $0, \mat{q}_2,\mat{q}_3,\ldots,\ldots$ is at least a fraction $1+\varepsilon$ of the volume of the simplex spanned by $0,\mat{e}_2,\mat{e}_3,\mat{e}_4,\ldots$.  Similarly, the volume of the simplex spanned by $0, \mat{q}_1,\mat{q}_2,\mat{q}_3,\mat{q}_4,\ldots$ is at least a fraction of the volume of the simplex spanned by $0, \mat{e}_1,\mat{e}_2,\mat{e}_3,\ldots$.  Observe that the volume of the simplex spanned by $0, \mat{q}_{1,2}, \mat{q}_3,\mat{q}_4,\ldots$ is a convex combination of the volumes of $\sigma$ and $\sigma'$, which implies that the height of $\mat{q}_{1,2}$ is at most a factor $1+\varepsilon$ less than the height of $\frac{1}{\sqrt{2}}\mat{e}_1 + \frac{1}{\sqrt{2}}\mat{e}_2$ from the affine subspace spanned by $\mat{e}_3,\mat{e}_4,\ldots$ (which is 1).  It follows that the sine of the angle between $\mat{q}_{1,2}$ and the affine subspace spanned by $\mat{q}_{3,4}, \mat{q}_{5,6}, \ldots$ is at least $1/(1+\varepsilon)$.  The same conclusion can be drawn about the sine of the angle between $\mat{q}_{3,4}$ and the affine subspace spanned by $\mat{q}_{1,2},\mat{q}_{5,6},\ldots$, and so on.

  Therefore, we can repeat the above argument $\log_2 |Q| \geq \log_2|P|-1$ times.  In the end, we produce a vector $\mat{q}$ such that the cosine of the angle between $\mat{h}$ and $\mat{q}$ is at least $(\sqrt{3/2})^{\log_2 |P|-1} \cos\theta \geq \sqrt{2/3} \cdot |P|^{(\log_2 3 - 1)/2}/N$.
\end{proof}

Initially $N = \log m$.  We take a $\log m$ disjoint subsets $P_i$ of $W = \{\mat{w}_i : i \in [\rho]\}$, each consisting of $\log m$ vectors. By Lemma~\ref{lem:tech}, we obtain a conical combination $\mat{u}_i$ of $P_i$ such that the cosine of the angle between $\mat{u}_i$ and $\mat{h}$ is at least $\sqrt{2/3}\cdot (\log m)^{(\log_2 3 - 1)/2}/\log m = \sqrt{2/3} \cdot (\log m)^{(\log_2 3 - 3)/2}$.

Imagine that we lay a very fine grid over the unit sphere centered at the origin in $\mathbb{R}^s$.  Given a subset $Y$ of $m$ grid points, we call a grid point \emph{realizable with respect to $Y$} if that grid point is a conical combination of points in $Y$.  For every subset of at most $\log^4 m$ realizable points with respect to $Y$, we use $Y_R$ to denote $Y \cup R$.  Consider the collection $\mathcal{Z} = \{Y_R\}$ over all possible subset $Y$ of $m$ grid points and all possible subset $R$ of $\log^4 m$ realizable grid points with respect to $Y$.  The collection $\mathcal{Z}$ is huge but finite.

When we apply the random projection $\varphi$ to $X$, we assume for simplicity that $\varphi(X)$ is equal to some $Y$ as the grid in $\mathbb{R}^s$ is very dense.  For a subset $R$ of at most $\log^4 m$ realizable points with respect to $Y$, the preimage of $R$ under $\varphi$ is the corresponding set conical combinations of $X$.  The preimage of $Y_R$ is just the preimage of $Y \cup R$.  Imagine that we apply $\varphi$ to the preimages of all $Y_R \in \mathcal{Z}$ under $\varphi$ simultaneously.  So $\varphi$ will produce the dimension reduction with low volume distortion on some preimages.  Without loss of generality, we can assume that this is the case with $X_R$, where $R$ includes the grid points closest to $\mat{u}_1,\mat{u}_2,\ldots,\mat{u}_{\log m}$, respectively.  Since the grid can be very fine, we just assume that $R$ contains $\mat{u}_1,\mat{u}_2,\ldots,\mat{u}_{\log m}$ for simplicity.

As a result, we can now apply Lemma~\ref{lem:tech} to $\{\mat{u}_1,\ldots,\mat{u}_{\log m}\}$ and conclude that they have a conical combination $\mat{u}'_1$ such that the cosine of the angle between $\mat{u}'_1$ and $\mat{h}$ is at least $(2/3) \cdot (\log m)^{(\log_2 3 -1)/2} \cdot (\log m)^{(\log_2 3 - 3)/2} = (2/3) \cdot (\log m)^{\log_2 3 - 2}$.  Similarly, we can draw subsets $P_i$ for $i \in [\log m + 1, 2\log m]$, produce $\mat{u}_{\log m + 1},\ldots, \mat{u}_{2\log m}$, and then produce $\mat{u}'_2$ such that the cosine of the angle between $\mat{u}'_2$ and $\mat{h}$ is at least $(2/3) \cdot (\log m)^{\log_2 3 - 2}$.  We can continue this way to reduce the exponent of $\log m$ to zero.  The result is that $\mat{w}_1,\ldots,\mat{w}_\rho$ has a conical combination that makes an angle no more than $\arccos(4/9)$ with $\mat{h}$.  We need $\rho = \Theta(\log^4 \nu)$ to carry out the above argument to reduce the exponent of $\log m$ to zero.

\end{document}